\newif\ifsubmission
\submissionfalse  %

\ifsubmission
\documentclass[sigconf,review]{acmart}
\else
\documentclass[sigconf,authordraft]{acmart}
\fi
\setcopyright{acmcopyright}
\copyrightyear{2019}
\acmYear{2019}
\acmDOI{}

\ifsubmission
\acmConference[ICSE '20]{ICSE '20: International Conference on Software Engineering}{May 23-29, 2020}{Seoul, South Korea}
\acmBooktitle{}
\acmISBN{0000}
\else
\acmConference[Preprint]{Preprint}{Aug 2019}{not for distribution.}
\acmBooktitle{Preprint}
\acmISBN{0000}
\fi

\usepackage{xspace}
\usepackage{xcolor}
\definecolor{diffincl}{RGB}{34,139,34} %
\definecolor{diffrem}{named}{red}
\usepackage{multirow}

\usepackage[utf8]{inputenc}

\title{Learning to Fix Build Errors with Graph2Diff Neural Networks}

\author{Daniel Tarlow}
\affiliation{
\institution{Google}
}

\author{Subhodeep Moitra}
\affiliation{
\institution{Google}
}

\author{Andrew Rice}
\affiliation{
\institution{University of Cambridge \\ \& Google}
}

\author{Zimin Chen}
\authornote{Work done during internship at Google.}
\affiliation{
\institution{KTH Royal Institute of Technology}
}

\author{Pierre-Antoine Manzagol}
\affiliation{
\institution{Google}
}

\author{Charles Sutton}
\affiliation{
\institution{Google}
}

\author{Edward Aftandilian}
\affiliation{
\institution{Google}
}

\usepackage{natbib}
\usepackage{graphicx}
\usepackage{pbox}
\usepackage{amssymb} %
\usepackage{subcaption}
\usepackage{amsmath}
\usepackage{stmaryrd}

\usepackage{listings}
\usepackage{array}
\usepackage[inline,shortlabels]{enumitem} %
\usepackage{verbatim}

\setitemize{noitemsep,topsep=0pt,parsep=0pt,partopsep=0pt,leftmargin=2pt}
\setenumerate{noitemsep,topsep=0pt,parsep=0pt,partopsep=0pt,leftmargin=2pt}

\newcolumntype{C}[1]{>{\centering\let\newline\\\arraybackslash\hspace{0pt}}m{#1}}

\lstnewenvironment{JavaDiff}[1][]{
\lstset{language=Java,
keywords={}, %
morecomment=[f][\color{diffincl}]{+\ },
morecomment=[f][\color{diffrem}]{-\ }, #1}}{}

\newcommand{\secref}[1]{Sec.~\ref{#1}}
\newcommand{\figref}[1]{Figure~\ref{#1}}
\newcommand{\lstref}[1]{Listing~\ref{#1}}

\newcommand{\diagnostics}[1]{\footnotesize{\texttt{#1}}}
\newcommand{\deepdelta}{\textsc{DeepDelta}\xspace}

\newcommand{\denotation}[2]{\ensuremath{\llbracket #1 \rrbracket_{#2}}}

\newcommand{\calL}{\mathcal{L}}
\newcommand{\bfs}{\mathbf{s}}
\newcommand{\topequiv}{\Leftrightarrow}
\newcommand{\concat}[1]{\ensuremath{\mathrm{concat}(#1)}}

\ifsubmission\else
\newcommand{\mainsecref}[1]{Sec.~\ref{#1}}

\newcommand{\maineqref}[1]{\eqref{#1}}
\fi

\begin{document}

\begin{abstract}
Professional software developers spend a significant
amount of time fixing builds, but this
has received little attention as a problem in automatic
program repair.
We present a new deep learning architecture, called Graph2Diff, for automatically localizing and fixing 
build errors. We represent  source code, build configuration files,
and compiler diagnostic messages as a graph,
and then use a Graph Neural Network model to predict a diff.
A diff specifies how to modify the code's abstract syntax tree, represented in the neural network 
as a sequence of tokens and of pointers to
code locations.
Our network is an instance of
a more general abstraction which we call
Graph2Tocopo, which is potentially useful in
any development tool for predicting
source code changes.
We evaluate the model on a dataset of over 500k real build errors and their resolutions from
professional developers.
Compared to the approach of DeepDelta \cite{mesbah2019fixing}, our approach tackles the harder task of predicting a more precise diff but still achieves over double the accuracy.
\end{abstract}

\maketitle

\section{Introduction}

Professional software developers spend a significant amount of time fixing builds;
for example, one large-scale study found that developers build their code
7--10 times per day \cite{seo2014programmers}, with a significant number of builds 
being unsuccessful. 
Build errors include simple errors such as syntax errors,
but for professional developers these are a small minority of errors;
instead, the majority are linking errors such as unresolved symbols, type errors,
and incorrect build dependencies \cite{seo2014programmers}.
A recent paper by Google reports roughly 10 developer-months of effort are spent every month fixing small build errors~\cite{mesbah2019fixing}.
Therefore,
automatically repairing build errors is a research problem
that has potential to ease a frequent pain point in the
developer workflow.
Happily, there are good reasons to think that
automatic build repair is feasible: fixes are often short (see Section~\ref{sec:dataset}), and we can test a proposed
fix before showing it to a developer simply by rebuilding
the project.
Build repair is thus a potential ``sweet spot'' for automatic program
repair, hard enough to require new research ideas,
but still just within reach.

However, there is only a small literature on repairing build errors.
Previous work on syntax errors has been very successful generating repairs resolving missing delimiters and parentheses \citep{campbell14natural,gupta2017deepfix}.
In contrast, in our corpus of professional
build errors (Table~\ref{fig:example-fixes}), fixes
are more subtle, often requiring detailed
information about the project APIs and dependencies.
Recently, the \deepdelta system 
\cite{mesbah2019fixing} aimed to repair build errors by
applying neural machine translation (NMT), translating
the text of the diagnostic message to a description of the
repair in a custom domain-specific language (DSL).
Although this work is very promising, the use of
an off-the-shelf NMT system severely limits the types of build
errors that it can fix effectively.

To this end, we introduce a new deep learning architecture, called \emph{Graph2Diff networks}, 
specifically for the problem of predicting edits to source code, as a replacement for the celebrated
sequence-to-sequence model used for machine translation. 
Graph2Diff networks map a \emph{graph} representation of 
the broken code to a \emph{diff}\footnote{We slightly abuse terminology here and use ``diff'' to mean a sequence of edit operations that can be applied to the broken AST to obtain the fixed AST.} in a domain-specific language that describes the repair.
The diff can contain not only tokens, but also pointers
into the input graph (such as ``insert token HERE'')
and copy instructions (i.e. ``copy a token from HERE in the input graph'').
Thus, Graph2Diff networks combine, extend, and generalize a number of recent ideas from  neural network models for source code \citep{maddison2014structured,allamanis2016convolutional,allamanis2017learning,zhao2019neural}.

Graph2Diff networks are based on three key architectural ideas from
deep learning:
graph neural networks, pointer models, 
and copy mechanisms.
Each of these ideas addresses a key challenge in modelling source
code and in  program repair.
First, \emph{graph neural networks} \cite{scarselli2009graph,li2016gated} can explicitly encode syntactic structure, semantic information, 
and even information from program analysis
in a form that neural networks can
understand,
allowing the network to learn to change one part of the code based on its relationship to another part of the code.
Second,
\emph{pointer models} \cite{vinyals2015pointer} can generate locations in the initial AST to be edited, which leads to a compact way of generating changes to large files (as diffs).
Much work on program repair divides the problem into two separate
steps of fault localization and generating the repair;
unfortunately, fault localization is a difficult problem \cite{liu2019you}.
Using pointers, the machine learning component can predict
both where and how to fix.
Finally, the \emph{copy mechanism} addresses the
well-known \emph{out-of-vocabulary} problem of source code \cite{allamanis2013mining,hellendoorn2017fse,karampatsis2019maybe}: source code projects often include project-specific identifiers
that do not occur in the training set of a model. A copy mechanism
can learn to copy any needed identifiers from
the broken source code, even if the model has never encountered the
identifier in the training set. 
Essentially, the copy mechanism is a way to encode into a neural network the insight from prior research on program repair \citep{le2011genprog,barr2014plastic} that often code
 contains the seeds of its own repair. Copy mechanisms give the model a natural way to generate these fixes.
Graph2Diff networks are an
instantiation of a more general abstraction that we introduce, called Graph2Tocopo, which encapsulates the key ideas of graphs, copy mechanisms, and pointers into a simple conceptual
framework that is agnostic to the machine learning approach, i.e., not just deep learning.

Our contributions are:
\begin{enumerate}
\item We study the challenges in repairing build errors seen in production code drawn from our collected dataset of $500,000$ build repairs (\secref{sec:dataset}). These observations motivate the requirements for a build repair tool: source code context is required, disjoint but correlated changes are often required, and repairs do not always take place at diagnostic locations.

\item We introduce the \emph{Graph2Tocopo} abstraction (\secref{sec:graph2tocopo}) and \emph{Graph2Diff network} for predicting source
code edits (\secref{sec:graph2diff}).
They are particularly well-suited to code-to-edit problems and have desirable properties relative to Sequence-to-Sequence models and other Graph-to-Sequence models. In this paper we show the value of this architecture for build repair, but in general this formulation is relevant to other tasks which require predicting changes to code.
\item Based on an \emph{extensive evaluation}
of our large historical data set of build errors,
we find that the Graph2Diff networks have remarkable performance,
achieving a precision of 61\% at producing the exact developer fix when suggesting
fixes for 46\% of the errors in our data set.
They also achieve over double the accuracy of the state-of-the-art
\deepdelta system.
Finally, we show that in some cases where the proposed fix does not match the developer's fix, the proposed fix is actually preferable.
\end{enumerate}

Overall, our results suggest that
incorporating the syntactic
and semantic structure of code
has a significant benefit in 
conjunction with deep
learning.
In future work, we hope that our
work provides a framework for
enhancing deep learning with more
sophisticated semantic information from
programs, ranging from
types to program analysis.

\section{Problem Formulation}

Here we formulate the problem of resolving a set of build diagnostics.
The input is the state of the source tree at the time of a broken build and a list of diagnostics returned by the compiler.
The target output is a diff that can be applied to the code to resolve all of the diagnostics.
For our purposes, a diff is a sequence of transformations to apply to the original source code to generate the repaired version.
Compiler diagnostics do not always identify the source of the fault that needs to be repaired, so we require the models to predict the locations that need changing in addition to the repairs.
This combines the well-studied problems of automated fault localization and automated program repair.

\subsection{Input data format}
\label{sec:input_data_format}

We represent source code files as Abstract Syntax Trees (ASTs) in order to capture the syntactic structure of the code (e.g., a statement is within a block that is within a for loop within a method declaration within a class, etc). Following \deepdelta \cite{mesbah2019fixing}, we also parse build configuration files into a similar structure.

Build errors are represented as a set of compiler diagnostics. A compiler diagnostic includes the kind of diagnostic (e.g., compiler.err.cant.resolve), the text associated with the diagnostic (e.g., ``Cannot resolve symbol \texttt{WidgetMaker}''), and a location that is composed of a filename and line number. We further assume that the diagnostic text can be decomposed into a text template and a list of arguments (e.g., template ``Cannot resolve symbol'' and arguments list [``\texttt{WidgetMaker}'']).

\subsection{Output data format}

The target output is a sequence of edits that can be applied to the ASTs of the initial ``broken'' code in order to resolve all the diagnostics (thus producing the ``fixed'' ASTs). In general, a resolution may require changing multiple files, but in this paper we restrict attention to fixes that only require changing a single file. To enable the use of ASTs, we also discard broken code that cannot be parsed, not counting these cases in the results for this paper.

We use the GumTree code differencing tool \citep{falleri2014fine} to compute the difference between the broken AST and the fixed AST. We convert the tree differences into an edit script, which is a sequence of insertion, deletion, move, and update operations that can be applied to the broken AST to produce the fixed AST.
There is a question of how to represent edit scripts so that they can most easily be generated by machine learning models.
In \secref{sec:graph2tocopo} we present a general abstraction for representing edit scripts,
and in \secref{sec:fixing_build_errors} we present the specific instantiation that we use for fixing build errors.

\subsection{Problem Statement}

We can now state the problem that we focus on. Given a set of build diagnostics, the AST of the file that needs to be changed, and the associated BUILD file AST, generate the edit script that the developer applied to resolve the diagnostics.

We view the problem as the second, core stage of a two-stage prediction process.
Stage 1 predicts the file that needs to be changed from the build diagnostics. Stage 2 uses the result of Stage 1 to construct the AST of the file that needs to be changed and an AST of the associated BUILD file. The Stage 2 problem is then the problem statement above. Because Stage 2 is the core challenge that we are interested in, we use a heuristic for Stage 1 of choosing to edit the file with the most diagnostics associated with it,
and we limit our experiments to examples where the Stage 1 prediction is correct ($\sim$90\% in our data).
In practice if one were deploying such a system, the 10\% of cases where the Stage 1 prediction is incorrect should be treated as errors.

\section{Dataset}
\label{sec:dataset}

Before describing our approach, we describe the data in more detail, as it motivates some of the modeling decisions. 
\ifsubmission
\else
To collect data, we take advantage of the highly-instrumented development process at Google \citep{potvin2016google}, extending previous work by \cite{mesbah2019fixing}.
\fi

\subsection{Build Repair Dataset}

\begin{figure}[tb]
\centering
\begin{tabular}{ccc}
\hspace{-12pt}
\includegraphics[width=.345\columnwidth]{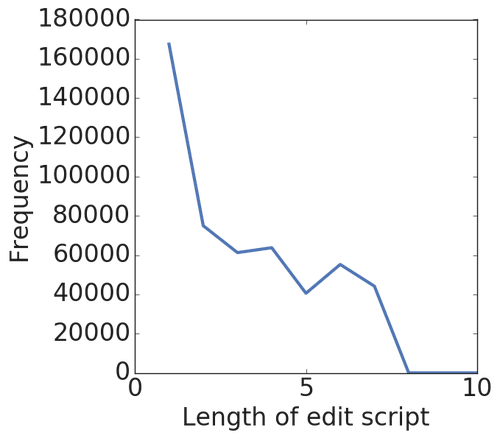} &
\hspace{-12pt}
\includegraphics[width=.34\columnwidth]{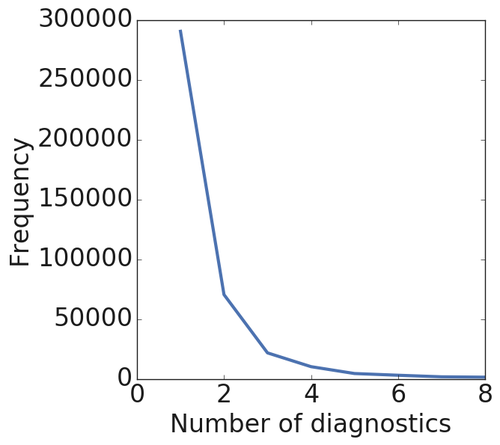} &
\hspace{-12pt}
\includegraphics[width=.32\columnwidth]{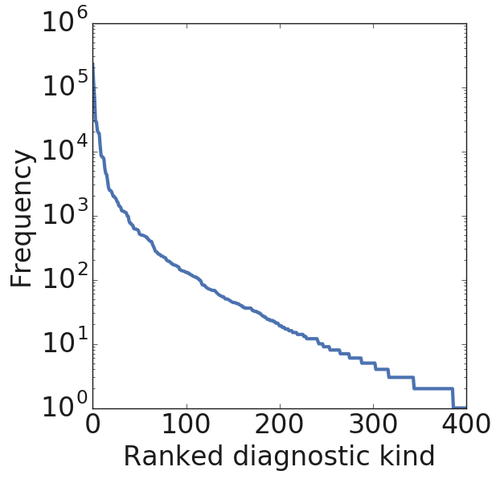} \\
(a) Fix length & (b) \# diagnostics & (c) Diagnostic freq.
\end{tabular}
\caption{Quantitative data statistics.\label{fig:data_statistics}}
\end{figure}

We collected data by looking at one year of build logs and finding successive build attempts where the first build produced compiler diagnostics and the second resulted in no diagnostics. We then retrieved the associated ``broken'' and ``fixed'' snapshots of the code.
These are examples where a developer repaired one or more build errors. We limit attention to examples with edit scripts of 7 or fewer edit operations, where the broken snapshot can be parsed successfully, and that fix the build by changing the Java code of the file with the most number of diagnostics (i.e., discard fixes that only change build configuration files, command line flags, and unexpected files).
We do not restrict the kinds of diagnostics, the vocabulary used in the edit scripts, or the sizes of input files.
The result is a dataset of $\sim$500k fixes.
\figref{fig:example-fixes} shows six examples from the dataset (with identifiers renamed) and \figref{fig:data_statistics} shows quantitative statistics.
We make the following observations:

\begin{figure*}[t]
\begin{minipage}{\textwidth}
\begin{tabular}{|c | C{.33\textwidth} |l|}
\hline
{\bf Row} & {\bf Diagnostics} & {\bf Fix} \\
\hline
\hline 
A 
&
\diagnostics{cannot find symbol `widgetSet()'}
& 
\begin{JavaDiff}
-    widgetsForX.widgetSet().stream().forEach(
+    widgetCounts.widgetSet().stream().forEach(
\end{JavaDiff}
\\ \hline
B 
&
\diagnostics{cannot find symbol `of'}
&
\begin{JavaDiff}
-    Framework.createWidget(new FooModule.of(this)).add(this);
+    Framework.createWidget(FooModule.of(this)).add(this);
\end{JavaDiff}
\\ \hline
C
&
\diagnostics{(line 10) cannot find symbol `longName' \newline
(line 15) cannot find symbol `longName'}
&
\begin{JavaDiff}
-    String longname = "reallylongstringabcdefghijklmnopqrstuvw..."
+    String longName = "reallylongstringabcdefghijklmnopqrstuvw..."
     x = f(longName)  // (line 10)  Diagnostic pointed here
     // ...
     y = g(a, b, c, longName) // (line 15)  Diagnostic pointed here
\end{JavaDiff}
\\ \hline
D 
&
\diagnostics{incompatible types: ParamsBuilder cannot be converted to Optional<Params>}
&
\begin{JavaDiff}
-    return new ParamsBuilder(args);
+    return new ParamsBuilder(args).build();
\end{JavaDiff}
\\ \hline
E
&
\diagnostics{incompatible types: RpcFuture<LongNameResponse> cannot be converted to LongNameResponse}
&
\begin{JavaDiff}
-  LongNameResponse produceLongNameResponseFromX(
+  ListenableFuture<LongNameResponse> produceLongNameResponseFromX(
\end{JavaDiff}
\\ \hline
F
&
\diagnostics{incompatible types: WidgetUnit cannot be converted to Long}
&
\begin{JavaDiff}
-  public Widget setDefaultWidgetUnit(WidgetUnit defaultUnit) {
+  public Widget setDefaultWidgetUnit(Long defaultUnit) {
     this.defaultUnit = defaultUnit;    // Diagnostic pointed here
\end{JavaDiff}
\\ \hline
G
&
\diagnostics{cannot find symbol `of(Widget,Widget)'}
&
\begin{JavaDiff}
- import com.google.common.collect.ImmutableCollection;
+ import com.google.common.collect.ImmutableSet;
// ...
-            ImmutableCollection.of(
+            ImmutableSet.of(
\end{JavaDiff}
\\ \hline
\end{tabular}
\end{minipage}
\caption{Example diagnostics and fixes from our dataset.}
\label{fig:example-fixes}
\end{figure*}

{\bf Variable Misuse~\citep{allamanis2017learning} errors occur.} Row A shows one of 6\% of the cases where the fix replaces a single wrong usage of a variable.
    
{\bf Source-code context is required since the same diagnostic kind has many different resolutions.} Rows A-C are \texttt{cant.resolve} diagnostics, and rows D-F are \texttt{incompatible.types} diagnostics. Each requires a different replacement pattern. \figref{fig:data_statistics}(c) shows the frequency of diagnostic kinds in the dataset. A small number of diagnostic kinds dominate, but the graph has a heavy tail and there are numerous resolution patterns per diagnostic, which means that a learning-based solution (as opposed to attempting to build a hand crafted tool for these diagnostics) seems a good option.
    
{\bf Edit scripts can be relatively long.} Row E requires an edit script of length 4; by comparison, a Variable Misuse bug such as in Row A can be fixed with an edit script of length 1. \figref{fig:data_statistics}(a) shows the overall distribution of edit script lengths.
	
{\bf Fixes do not always occur at the diagnostic location.} Row A shows an example where the identifier in the diagnostic is not the one that needs changing. Rows C and F show examples where the diagnostics indicate a different line to the one that needs changing. 36\% of cases require changing a line not pointed to by a diagnostic.
	
{\bf There can be multiple diagnostics.} Row C shows an example where there are multiple diagnostics. \figref{fig:data_statistics}(b) shows the distribution of diagnostic frequency per build. Approximately 25\% of failures had more than one diagnostic.
	
{\bf Single fixes can span multiple locations.} In Row G, multiple code locations need to be changed in order to fix the error. 
The changes at the different locations are part of a single fix but require different changes at the different locations. This shows that we need a flexible model which does not assume that fixes at different locations are independent as in DeepFix \citep{gupta2017deepfix}, or that multi-hunk fixes apply the same change at different locations as in \cite{saha2019harnessing}.
21\% of the data requires editing more than one contiguous region.

\section{Graph2Tocopo Abstraction}
\label{sec:graph2tocopo}

Our aim is to develop a machine learning approach that can handle all of the complexities described in the previous section.
These challenges appear not only in fixing build errors, but also in many other code editing tasks.
To match the generality of the challenges, we start by developing an abstraction for code editing called \emph{Graph2Tocopo}.
Graph2Tocopo aims to formalize the interface between 
the machine learning method and the code-editing problem --- like 
program repair, auto-completion, refactoring, etc.
Graph2Tocopo is a single abstraction that unifies ideas
from many recent works in modelling source code \citep{maddison2014structured,allamanis2016convolutional,bhoopchand2016learning,allamanis2017learning,vasic2019neural,zhao2019neural}. 
Though we will take a deep learning approach in \secref{sec:graph2diff},
Graph2Tocopo is \emph{not} specific to deep learning.

Graph2Tocopo aims to crystallize three key concepts that 
recur across code editing tasks: representing code as graphs,
representing pointers to code elements, and
copying names from code.
Representation of code as a graph gives a convenient way of
representing code abstractly; of combining multiple sources of information, such as code, error messages, 
documentation, historical revision information, and so on;
and for integrating statistical, syntactic, and semantic structure by constructing edges.
Graphs are a natural choice because they are already a lingua
franca for (non-learning based) syntactic and semantic analysis of code.
At a high level, the goal of the Graph2Tocopo abstraction
is to do for code-editing tasks what the celebrated Seq2Seq abstraction \citep{sutskever2014sequence}
 has done for natural language processing (NLP):
Graph2Tocopo aims
to serve as an interface between \emph{tool developers}, that is, software
engineering researchers who create new development tools
based on predicting code edits,
and \emph{model developers},
machine learning researchers developing new
model architectures and training methods.
Part of our goal is to encourage modularity between
these two research areas, so advances one on side
can immediately benefit the other.

When designing a tool
for a particular code-editing task,
we envision that the tool developer will develop two
formalisms, one for the input and one for the output.
On the input side, the tool developer
designs 
a graph to represent
the program to be edited,  abstracting the code in
a way that reveals the most
important information for the task (\secref{subsec:tocopo_graph}).
On the output side, the tool
developer develops 
an \emph{edit-domain specific language (eDSL)} that formalizes the class of code edits that is necessary for the tool.
Statements in the eDSL are  \emph{Tocopo sequences} (\secref{subsec:tocopo_tocopo}), that is, sequences that contain either tokens or
two different types of location references,
called copy operations and pointers.
Afterwards, we envision that the tool developer 
could choose between many different learning methods for
Graph2Tocopo (such as \secref{sec:graph2diff}), 
without going into the details of how the models are implemented. 

\subsection{Code as Graphs}\label{subsec:tocopo_graph}

We represent code and related context as a directed multi-graph with discrete labels on nodes and edges.
For different tasks, the graph can include different information such as abstract syntax trees, error messages, results of program analyses, and edges relating these components, e.g., a diagnostic line number refers to a location in the code. 
Each node is specified by an integer index $i$ and is associated with a tuple of node features $(t_i, v_i)$, where $t_i$ is a member of a finite set of node types $\mathcal{T}$ and $v_i$ is a string called the \emph{node value}.
E.g.,~to represent ASTs, $\mathcal{T}$ can be the set of nonterminals used by Java, and node values $v_i$ could represent literals, keywords, and identifiers.
Edges are specified by a triple $(i, j, e)$, which means that there is an edge from $i \rightarrow j$ with type $e$.

\subsection{Edits as Tocopo Sequences}\label{subsec:tocopo_tocopo}

The second challenge is how to represent a code edit.
We propose to use a sequence of code locations and tokens that we call a \emph{Tocopo sequence}.
In more detail, a Tocopo sequence is a sequence of Tocopo expressions, where 
a Tocopo expression is one of
(a) a \emph{token expression} of the form \texttt{TOKEN($t$)}.
These represent literal tokens, which could be commonly occurring
identifier names, or editing commands like \texttt{INSERT} and \texttt{UPDATE},
(b) a \emph{copy expression}, which 
refers to a value
in the input graph
(c) an \emph{input pointer expression},
which refers to a specific node in the input graph, and
(d) an \emph{output pointer expression} which refers
to a previous element in the Tocopo sequence.
This syntax is given in
\figref{fig:tocopo_sequence_syntax}.

We assume that a tool designer creates an eDSL to represent edits to source code in a way that is appropriate for the task at hand.
The tool designer chooses a set of keywords for the edit DSL and how they
are combined with code locations and code tokens to represent an edit.
Just as a programming language is a subset
of the set of all sequences of tokens, an edit DSL is as a subset
of the set of all Tocopo sequences,
that is, sequences of tokens, copy, and pointer operations.
These three types of expressions
are useful for constructing eDSLs for
a variety of code editing tasks.

\begin{figure}
\begin{minipage}{\columnwidth}
\fbox{
    \begin{tabular}{lcl}
        $\boldsymbol{s} ::= s_1, \ldots, s_M$ && \hbox{Tocopo sequence} \\
        $s ::=$  && \hbox{Tocopo expression} \\
        \qquad $\texttt{TOKEN}(t)$ && \hbox{\qquad Token expression} \\
        \qquad $\texttt{COPY}(n)$ && \hbox{\qquad Copy expression} \\
        \qquad $\texttt{INPUT\_POINTER}(n)$ && \hbox{\qquad Input pointer expression} \\
        \qquad $\texttt{OUTPUT\_POINTER}(m)$ && \hbox{\qquad Output pointer expression} \\
    \end{tabular}
}
\caption{Syntax of Tocopo sequences. Here
$t$ denotes a token, and
$n$ and $m$ integers.} \label{fig:tocopo_sequence_syntax}
\end{minipage}
\end{figure}

The concept of Tocopo sequence is extremely generic, and does
not say anything about what the edits do.
What we can say about Tocopo sequences 
at this general level is what the references
mean, that is, the pointers and the copy operations.
Given a graph $G,$
a token expression $\texttt{TOKEN}(t)$ can be interpreted
simply as referring to $t.$
A copy expression $\texttt{COPY}(n)$ refers to the value $v_n$
of node $n.$
An input pointer expression
$\texttt{INPUT\_POINTER}(n)$ refers to node index $n$
in $G.$
 Finally, given a Tocopo sequence $s_1 \ldots s_M$,
$\texttt{OUTPUT\_POINTER}(j)$ for $j < M$ refers to $s_j.$ 
This allows us to define two Tocopo sequences
 $s_1 \ldots s_M$ and $s'_1 \ldots s'_M$
to be referentially equivalent,
which we write $\bfs \topequiv_{{G}} \bfs'$,
if for all $i \leq M$, the two expression $s_i$ and $s_i'$ refer
to the same entity.
In practice the equivalence arises when token $t$ referred to in one
expression is equivalent to the node value referred to by a copy operation in the other expression.
A key constraint on the design of eDSLs, which we will leverage
in the learning algorithm (\secref{sec:graph2tocopo}) is
that if $\bfs' \topequiv_{{G}} \bfs$, then $\bfs$ and $\bfs'$
specify the same edit to $G.$

\emph{Remarks.} First, it may be unclear why we need output pointers. 
These are useful for adding new larger subtrees to code, because
using output pointers we can add multiple new nodes to the AST with a
desired relationship among them.
Second, the distinction between pointers and copy
operations is subtle, but important.
Sometimes, it
is important to specify an exact code \emph{location},
such as when specifying where a new AST node
should be inserted. This is a pointer.
Other times, it is useful to refer to the \emph{value} of an input node,
such as when inserting a usage of a previously-defined variable, but any
other location with the same node value will do just as well.
This is a copy.
Essentially, this is the code-editing version of the classic distinction between pass-by-value and pass-by-reference in programming  languages.

\subsection{Program Repair as Graph2Tocopo}

To be more concrete,
we give examples of several program
repair methods from the literature, showing how they can be  
represented as Graph2Tocopo problems. 
(An example of a build repair specified as a Tocopo sequence is shown in \figref{fig:example_edit_script}.)
First, \citet{allamanis2017learning} propose learning to repair variable misuse errors,
such as the one in \figref{fig:example-fixes}A, using a graph neural network.
They proposed using a graph called an \emph{augmented AST},
that is, the AST of the program with additional edges to represent
adjacent tokens, data flow, and control flow. 
Additionally, in each program, there is one identifier expression 
for which
we know a previously-declared variable should be used, but we do not know which one.
This expression is represented by a special node in the graph of type \texttt{HOLE},
which has child nodes of type \texttt{CANDIDATE} 
that represent all of the variables in scope.
The goal is to predict which of the candidate nodes is the correct
variable.
This task can represented as a Graph2Tocopo problem in a very simple way.
The input graph is simply the augmented AST,
and the output Tocopo sequence has the form
\texttt{INPUT\_POINTER($j$),} where $j$ is the node ID of one of the candidate nodes in the graph.

\citet{vasic2019neural} suggest a method for jointly localizing and repairing variable misuse errors, using two pointers: the first is a reference to the variable usage that is an error
(or to a special sequence position that indicates no error), and the second pointer is a reference to another variable that should be used in place of the first.
This can be represented as a Tocopo sequence of
$\texttt{INPUT\_POINTER(i)} \,\, \texttt{COPY(j)}$,
where $i$ is the node id of the incorrect variable usage,
and $j$ is the node id of any usage of the correct replacement.
Note the difference between pointers  and copy operations.
A pointer is necessary for
the first usage, while a copy provides more flexibility
for the second.

\subsection{Learning for Graph2Tocopo}
\label{sec:graph2tocopo_learning}

Combining the graph-structured input with the Tocopo-structured output results in the Graph2Tocopo abstraction. 
A training set is collected of code snapshots represented as graphs,
and target edits represented as Tocopo sequences in the eDSL in question.
We can then treat learning as a supervised learning problem to map graphs
to Tocopo sequences.
A variety of learning algorithms can be applied to this task.
Many learning methods, especially in deep learning,
can define a probability distribution $p(\bfs | G, w)$
over Tocopo sequences $\bfs$ given graphs $G$
and learnable weights $w$. (For now, we treat
this distribution as a black box; see \secref{sec:graph2diff}
for how it is defined in the build repair application.)
Then, given a training set $D = \{({G}, \bfs)\}$
of graphs and target Tocopo sequences, the learning
algorithm chooses weights to maximize the probability
of the data, that is, to maximize the objective function
$\mathcal{L}(w) = \sum_{(G, \bfs) \in D} \log p(\bfs | G, w).$

For Graph2Tocopo sequences, though, we can do better than this standard approach,
in a way that eases the burden on the tool designer to choose which of potentially many reference-equivalent sequences should be provided as target output.
Consider a single example $({G}, \bfs)$ from the training set.
The tool designer should not worry about which equivalent sequence is desired when they all correspond to the same edit.
Thus, we recommend training the model to maximize the probability assigned to the \emph{set} of expressions equivalent to $\bfs$. That is,
let $\mathcal{I}_{{G}}(\bfs) = \{ \bfs' \,|\, \bfs' \topequiv_{{G}} \bfs \}$ be the set of equivalent expressions, and train using the objective function 
\begin{align}
    \mathcal{L}(w) 
    & = \sum_{(G,\bfs) \in D} \log \sum_{\boldsymbol{s}' \in \mathcal{I}_{{G}}(\boldsymbol{s})}
    p(\boldsymbol{s}' \mid {G}, w).
    \label{eq:marginal_likelihood}
\end{align}
This rewards a model for producing any Tocopo sequence that is reference-equivalent to the provided target sequence,
i.e., the model is free to use the copy mechanism as it sees fit.
It might seem that computing the objective function \eqref{eq:marginal_likelihood} is computationally intractable, as it may involve a sum over many sequences.
However, it can often be computed efficiently, and Graph2Diff models are constrained so that it becomes inexpensive to compute.

\section{Build Repair as Graph2Tocopo}
\label{sec:fixing_build_errors}

Now we cast build repair as a Graph2Tocopo problem.

\subsection{Input Graph Representation}

The input graph is composed of several subgraphs:

\emph{Code Subgraph.} We roughly follow \cite{allamanis2017learning} to represent source code as a graph, 
creating nodes for each node in the AST.
For identifiers and literals, the node value is the string representation that appears in the source code text. For internal nodes in the AST, the node value is a string rendering of the node kind. The node type is the kind of AST node as determined by the Java compiler.

\emph{Diagnostic Subgraphs.} There is one diagnostic subgraph for each compiler diagnostic. 
Nodes in this subgraph come from four sources.
First, there is a node representing the diagnostic kind as reported by the compiler, for example, \texttt{compiler.err.cant.resolve}.
Second, the text of the diagnostic message is tokenized into a sequence of tokens, each of which is added as a node in the graph.
Third, there is one node for each diagnostic argument (see \secref{sec:input_data_format}) from the parsed diagnostic message.
Finally, there is a diagnostic root node.
The subgraph has a backbone tree structure where the root node is a parent of each other listed node, and the nodes are ordered as above. 
For purposes of creating edges, we treat this tree as an AST.

\emph{BUILD File Subgraph.} BUILD file are usually an XML-style document (e.g.,~BUILD file in Bazel, POM file in Maven, build.xml in Ant), which we 
encode as a tree.

The subgraphs are connected by several types of edges, and we are planning to add more edge types. An ablation study that removes all edges (Supplementary Materials) in the input graphs shows the importance of these edges. Currently, we have:
\begin{itemize}
\item \textbf{AST child:} Connects all parents to their children.
\item \textbf{Next node:} Connects to the next node in a depth-first traversal.
\item \textbf{Next lexical use:} Connects nodes with a given value in a chain structure, with neighbors in the chain being nearest uses when nodes are ordered by the depth-first traversal. %
\item \textbf{Diagnostic location:} Connects each diagnostic root node to all nodes in the code subgraph on the line where the error occurred.
\item \textbf{Diagnostic argument:} Connects diagnostic argument nodes to the corresponding nodes in the code subgraph where its string value is equivalent to the diagnostic argument. 
\end{itemize}

\subsection{Output eDSL Design}

Here we describe the eDSL that we use for representing repairs to
build errors. A program in our eDSL, which we call an \emph{edit script},
specifies how to transform the broken AST into the fixed AST.
Our goals in designing the eDSL are (a) given two ASTs for
a broken file and a manually repaired file (which is what we have in our data),
it is easy to generate a corresponding edit script, and (b) edit scripts
should fully specify the change to the AST.\footnote{As natural as 
requirement (b) sounds, it is not always respected in previous work.} 
An edit script is a sequence of edit operations. Each edit operation specifies
one change to be made to the broken AST. 
As shorthand, we write \texttt{TOKEN}($t$) as $t$ and use \texttt{POINTER} to mean
that either an \texttt{INPUT\_POINTER} or \texttt{OUTPUT\_POINTER} is valid:
\begin{itemize}
\item \texttt{INSERT POINTER(<parentId>) POINTER(<siblingId>) <type> <value>}.
Inserts a new node of node type \texttt{type} 
and value \texttt{value} into the AST as a child of the node specified by parentId.
It is inserted into the children list after the referenced previousSibling node.
If the new node should be the first sibling,
then the special \texttt{FIRST\_CHILD} token is used in place of \texttt{POINTER(<siblingId>)}.
\item \texttt{DELETE INPUT\_POINTER(<nodeId>)} deletes a node.
\item \texttt{UPDATE INPUT\_POINTER(<nodeId>) <value>} sets the value of the referenced node to \texttt{value}.
\item \texttt{MOVE INPUT\_POINTER(<sourceId>) POINTER(<newParentId>) POINTER(<newSiblingId>)} moves the subtree rooted at 
\texttt{source} so that it is a child of \texttt{newParent}, occuring just after \texttt{newSibling}.
\item \texttt{DONE} indicates the end of the edit script.
\end{itemize}

For example, \figref{fig:example_edit_script} shows an edit
script that implements the fix from \figref{fig:example-fixes}E. 
Each operation in the edit script
adds one node of a three-node subtree in the AST that specifies
a Java parameterized type to be inserted, and then the old Java type is deleted.
This uses
input pointers, output pointers,
and values.

\begin{figure*}
\centering
\vspace{-5pt}
\begin{minipage}{0.75\textwidth}
\begin{JavaDiff}
-  LongNameResponse produceLongNameResponseFromX(
+  ListenableFuture<LongNameResponse> produceLongNameResponseFromX(
\end{JavaDiff}
\end{minipage}

\begin{minipage}{0.3\textwidth}
{\footnotesize\tt
\begin{tabbing}
0: MethodDef \\
1: \hspace{3pt} \= Type \\
2: \> \hspace{3pt} \=  LongNameResponse \\
3: \> Id \\
4: \> \> produceLongNameResponseFormX \\
5: \> Args \\
\end{tabbing}
}
\end{minipage}
\hspace{2em}%
\begin{minipage}{0.5\textwidth}
{\footnotesize
\begin{verbatim}
0:INSERT 1:INPUT_POINTER(1) 2:INPUT_POINTER(2) 3:PARAMETERIZED_TYPE 4:TYPEAPPLY
5:INSERT 6:OUTPUT_POINTER(0) 7:FIRST_CHILD 8:IDENTIFIER 9:ListenableFuture
10:INSERT 11:OUTPUT_POINTER(0) 12:OUTPUT_POINTER(5) 13:IDENTIFIER 14:LongNameResponse
15:DELETE 16:INPUT_POINTER(2)
17:DONE
\end{verbatim}
} 
\end{minipage}
\vspace{-10pt}
\caption{An example edit script that makes the change specified in Figure 1E. (Top) a textual diff of the change, (Left) A subset
of the original AST annotated with input ids.
(Right) An edit script implementing the change, annotated with output ids.}
\label{fig:example_edit_script}
\end{figure*}

\section{Graph2Diff Architecture}
\label{sec:graph2diff}

Finally we are able to describe our new deep learning architecture for Graph2Tocopo problems.
This architecture, which we call Graph2Diff has two components.
The first is a \emph{graph encoder} that converts the input graph into a $N \times H$ matrix called the \emph{node states},
where $N$ is the number of nodes in the input graph, and $H$ is the dimension of the hidden states in the network.
Each row of the node states matrix is a vector, which we call a \emph{node representation}, that corresponds to one node in the input graph,
and represents all the information about that node that the model has learned might be useful in predicting a fix.
The second component of Graph2Diff is an \emph{edit-script decoder} that predicts an edit script one Tocopo expression at a time, taking the node states and previous predictions as input.
The decoder is based on modern deep learning ideas for sequences, namely the celebrated Transformer architecture \cite{vaswani2017attention} which is used in models like GPT-2 \cite{radford2019language} and BERT \cite{devlin2019bert}, but requires custom modifications to handle the special Tocopo features of input pointers, output pointers, and copy operations.
Due to space constraints, we provide only a high-level
description here.
Full details are in the Supplementary Materials.

\subsection{Graph Encoder}

Inspired by \cite{allamanis2017learning}, we use a Gated Graph Neural Network (GGNN) to encode the graph into a matrix of node states. 
At a high level, a GGNN consists of a series of propagation steps.
At each step, a new representation of each node is computed by
passing the representations of the node's neighbors at the last step
through a neural network.
To initialize the node representations, we use a learnable continuous embedding for node's type and value, summing them together with a positional embedding \citep{vaswani2017attention} based on the order in the depth-first traversal.
We run GGNN propagation for a fixed number of steps (see \secref{sec:experiment_details} for details), resulting in
a representation for each node in the graph.

\subsection{Edit-Script Decoder}

The decoder is a neural network that predicts the edit script one Tocopo expression at a time.
If $V$ is the size of the vocabulary, a Tocopo expression
is either one of $V$ token expressions, one of $N$ input pointer
expressions, or one of $N$ copy expressions. So we can treat
predicting the next Tocopo expression as a classification
problem with $V + 2N$ outputs and predict this with a neural
network.\footnote{Our current implementation of the decoder handles output pointers in a simplified
way, predicting only where output pointers occur,
but not predicting what they point to. Therefore,
we treat \texttt{OUTPUT\_POINTER} as a vocabulary
item in this discussion.}
The inputs to the decoder are
(a) the node representations from the graph encoder, and (b)
a representation of the partial edit script generated so far.
Our decoder then builds on the decoder from the Transformer model, which is based on a
type of neural network called an \emph{attention operation}. An attention operation is a network that
updates the representation of a \emph{target sequence}
of length $N_2$, represented as a $N_2 \times H$ matrix, 
based on information from a \emph{source sequence} of length $N_1,$ represented
as a $N_1 \times H$ matrix.
The attention operation
produces an updated  $N_2 \times H$ matrix representing
the target. For
mathematical details, see our Supplementary Material.

Our edit-script decoder extends the Transformer
decoder to handle the pointer and copy
operations of Tocopo. Two main extensions are needed.
First, the partial edit script contains
Tocopo expressions, not just tokens as in Transformer,
so we need a way of representing
Tocopo expressions as vectors that can be used
within a deep network. 
To do this, we start with an ``output embedding'' step,
which produces a $T\times H$ matrix
of hidden states, where $T$ is the length of the partial edit script.
Then several layers of attention operations
alternate between (i)
exchanging information amongst the outputs,
via an attention operation where both the source
and target are the partial edit script
(known as ``causal self-attention''),
and (ii) sending information from the sequence
of nodes in the input graph to the output edit script,
via an attention operation where the source
is the input graph and the target is the partial
edit script
(which we call ``output-input attention'').
This results finally in a $T \times H$ matrix representing the partial edit script,
which is the input to an \emph{output layer},
which predicts the next expression.

This leads to the second extension. The output layer
 must produce a $V + 2N$ sized output of token, copy, and
pointer predictions
(whereas Transformer outputs just tokens).
To do this, our output layer makes three separate
predictions, which we call ``heads''.
The token head is an $H \times V$ matrix that maps the final hidden state
to a length $V$ vector of token scores.
The copy head and the pointer heads 
are both attention operations with different parameters
to produce length $N$ vectors of copy scores and pointer scores, respectively, as
in \cite{vinyals2015pointer}.
The three output vectors are concatenated into the $V + 2N$ outputs, and a softmax
is used to turn this into a distribution over predictions.

A final point is important but a bit technical.
In order to be able to efficiently train under the objective from
\eqref{eq:marginal_likelihood}, we require that the representation of
the Tocopo prefix provided to the decoder is the same for all
reference-equivalent prefixes; i.e., the network should make the same
future predictions regardless of whether previous predictions used
a token or an equivalent copy operation.
We impose this constraint by representing the partial edit script as
a list of \emph{sets} of all Tocopo expressions that are reference-equivalent
to each expression in the partial edit script.
These sets can be used within 
our attention operations with only minor modifications.
For details, see the Supplementary
Material.

\section{Experiments}
\label{sec:experiments}

We have several goals and research questions (RQs) for the empirical evaluation.
First, we would like to evaluate our design choices in Graph2Diff networks and better understand how they make use of available information.
We ask \textbf{RQ1: How is Graph2Diff performance affected by (a) the amount of code context in the input graph? (b) the model architectural choices? and (c) the amount of training data available?}
This question is important because the performance of deep learning methods can often be sensitive to model architectural choices.
To understand how our results fit in context with existing literature, we ask \textbf{RQ2: How do Graph2Diff networks compare to previous work on Seq2Seq for fixing build errors?}
We compare to the most closely related work, which is DeepDelta \cite{mesbah2019fixing}.
Even though DeepDelta can only be evaluated on a less-stringent task than exact developer fix match, we find that Graph2Diff networks achieve over double the accuracy, which shows that Graph2Diff networks
are far more accurate than previous work.
Turning attention to how the system would be used in practice, we ask \textbf{RQ3: How often do incorrect predictions build successfully?}, because fixes that fail to build can be filtered out and not presented to developers.
We find that 26\% $\pm$ 13\% of the incorrect predictions build successfully,
leading to an estimated precision of 61\% at producing the exact developer fix when suggesting
fixes for 46\% of the errors in our data set.
Finally, we ask the qualitative \textbf{RQ4: What kinds of fixes does the model get correct? What kinds of predictions are incorrect but build successfully?}
There are some cases where the fix is semantically incorrect and not desirable, but also cases where the predicted fix is preferable over the one provided by the developer.

\subsection{RQ1: Graph2Diff performance}

\subsubsection{Experimental details}
\label{sec:experiment_details}

We follow standard machine learning experimental details, using train, validation and test splits and grid search for choosing hyperparameters. Details appear in the Supplementary Materials.
Our main metric is sequence-level accuracy, which is how often the model predicts the full sequence correctly. This is a strict metric that only gives credit for exactly matching the developer's change. In the future we plan to show proposals to developers 
and measure how often they find them useful.

\subsubsection{Effect of context size and model depth}

It is possible to reduce the size of the input graphs by pruning nodes that are far away from the source of an error. Reducing graph sizes increases training throughput because less computation is needed for each example, and it may be possible for the learning problem to become easier if the removed nodes are irrelevant to the fix.

However, pruning nodes may also hurt performance of the model, for three reasons. First, if the error produced by the compiler is not near the source of the fault, then pruning can remove the location of code that needs to be changed and make it impossible for the model to correctly output the pointer elements in the output DSL. Second, the fix may require generating tokens that are out of vocabulary but present as the value of some input node. In this case, it is possible to generate the correct fix by outputting a copy element. However, if pruning removes all potential copy targets, then it will become impossible for the model to generate the correct answer. Third, there may be context in the distant nodes that are useful for the machine learning model in a less clear-cut way.

\begin{figure*}
    \centering
\begin{minipage}{.74 \textwidth}
\begin{tabular}{|c|c|c||c|c|c|}
 \hline
 \multirow{2}{*}{\parbox[b]{1.5cm}{\centering {\bf Prune\\Distance}}}  & & & \multicolumn{3}{c|}{{\bf Sequence-level Validation Accuracy}} \\
  & 
 {\bf Avg.~\# Nodes}     &
 {\bf \# Possible}     &
 {\bf 2 prop steps}     &
 {\bf 4 prop steps}  &
 {\bf 8 prop steps}  \\
     \hline
     \hline
 1    & 
 25     &
 125k     &
 14.9\%      &
 15.5\%   &
 15.7\%   \\
     \hline
 2&
41&
145k&
16.3\%&
16.9\%&
16.7\% \\
     \hline
4 &
192 &
240k &
19.9\% &
22.4\% &
23.4\%\\
     \hline
8 &
1524 &
310k &
23.8\% &
26.3\% &
{\bf 28.0\%} \\
     \hline
12 &
2385 &
315k &
22.8\% &
25.6\% &
27.1\% \\
     \hline
\end{tabular}
\end{minipage}%
\begin{minipage}{.21\textwidth}
\includegraphics[width=.9\columnwidth]{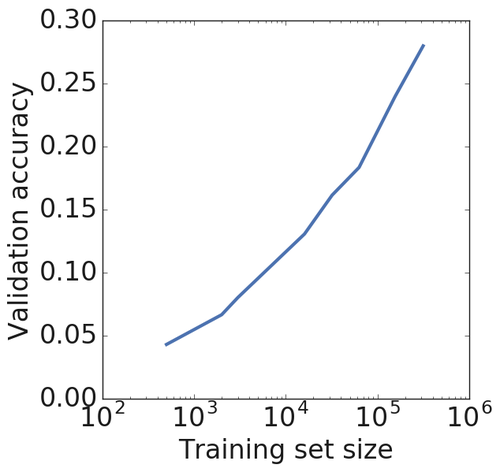}
\end{minipage}
\caption{(Left) Best sequence-level validation accuracy achieved for various degrees of graph pruning.
As the Prune Distance increases, more nodes are included in the graph, it becomes possible to get more training examples correct (the needed locations and vocabulary appear in the input graph), and accuracy generally increases.
More propagation steps leads to improved performance in most cases.
(Right) Best validation accuracy vs training set size.
}
    \label{fig:accuracy_vs_pruning}
\end{figure*}

Our first experiment explores this question. Results appear in \figref{fig:accuracy_vs_pruning} (left), showing that including more context and performing more propagation steps helps performance.

\subsubsection{Effect of dataset size}

To measure the effect of amount of data, we trained models using a random subsampling of the data. 
\figref{fig:accuracy_vs_pruning} (right)
shows the best resulting validation accuracy versus the number of training examples.
The x-axis is on a log-scale but clearly shows that increasing data size leads to improved performance.

\subsection{RQ2: Comparison to Sequence2Sequence}
\label{sec:deepdelta_experiments}

DeepDelta~\citep{mesbah2019fixing} casts the problem of fixing build errors in the Sequence2Sequence framework.
Here we compare Graph2Diff to DeepDelta across the two axes that they vary: Graph versus Sequence as input, and Diff versus Sequence as output.

\paragraph{Input: Graph vs Sequence. }
The main difference in the input representation is the amount of information provided to the models about the source code surrounding the error.
Within the context of Graph2Diff models, we can test how this choice affects performance while holding all else fixed.
To use the same input information as DeepDelta, we prune all nodes in the input graph except for the nodes on a path from the location of a diagnostic to the root of the AST.
We leave the diagnostic subgraph the same as in Graph2Diff models.
The result is a family of graph models that have a sequence input representation like that used in DeepDelta.
We call these models SeqGraph2X models, because they have sequential code input representations but are implemented within the Graph2Tocopo framework.
A benefit of the Graph2Tocopo framework is that they have a copy mechanism, unlike DeepDelta.

\paragraph{Output: Diff vs Sequence.}
Our diff output is more precise than the sequence output of DeepDelta in three ways:
(a) we refer to locations by pointing to nodes in the input graph, which resolves ambiguity when more than one input node has a given value (e.g., when changing a \texttt{private} modifier to \texttt{public} it becomes clear which \texttt{private} to change);
(b) we include a \emph{previous sibling} pointer to specify where we should insert into the list of children under the specified node, which resolves ambiguity about, e.g., order in argument lists;
(c) we generate AST types of new nodes to insert along with their value, which, e.g., resolves ambiguity between generating method invocations and method references.
The extra specificity in the diff output is important, because it provides enough information to automatically apply a fix generated by the model without additional heuristics or human judgement, which is crucial towards putting the system into practice.
Further, evaluating correctness in terms of matching an imprecise target output gives an overestimate of how the system will perform in practice.

The Graph2Tocopo framework makes it possible to run a series of experiments that gradually change the specificity of the output DSL from our precise diff output to the imprecise output from DeepDelta. We compare four output DSLs:
{\bf (1) ImpreciseDiff (Imprec)}: the output format from DeepDelta;
{\bf (2) ImpreciseWithPointers (Imprec+P)}: ImpreciseDiff but representing locations more precisely with pointers;
{\bf (3) ImpreciseWithPointersAndSiblings (Imprec+PS)}: ImpreciseWithPointers but adding previous sibling pointers; and
{\bf (4) Diff}: the Graph2Diff output DSL.

\paragraph{Graph2Diff vs DeepDelta. } Finally, we compare to a more direct reimplementation of DeepDelta, which uses the same sequential input representation but uses the Google Neural Machine Translation model \cite{wu2016google} for the Seq2Seq learning.
There is no pointer mechanism in this model, so it is not possible to evaluate on the more specific output 
DSLs and we compare just on ImpreciseDiff.
We equivalently refer to the DeepDelta method as Seq2ImpreciseDiff.

\paragraph{Experiment details and results}
We used the same experimental protocol as in the previous section to train the cross-product of options \{Seq, SeqGraph, Graph\} $\times$ \{Diff, ImpreciseWithPointersAndSiblings, ImpreciseWithPointers, ImpreciseDiff\}.
Accuracy is measured based on whether the model predicted the full sequence of its corresponding output correct (so generally predicting more abstract outputs is expected to produce higher accuracy).
We report differences in absolute performance compared to the Graph2Diff model on validation data.
Results appear in \figref{fig:seq_graph_to_abstract_diff_results}.

Comparing the first to the second row, we see that the SeqGraph2Tocopo formulation improves over the pure Seq2Seq formulation, which we attribute primarily to the copy mechanism that comes with the Tocopo-based output model. This is inline with other recent work that shows a benefit of a copy mechanism in program repair \cite{chen2018sequencer}.
Comparing the second row to the third, the graph-structured input improves performance regardless of the output DSL, and the importance of the graph grows as the specificity of the output DSL increases.
Also, as expected, performance increases as the output DSL becomes more abstract (but recall that we expect those other than Diff to overestimate real-world performance).
One other interesting comparison is Graph2ImpreciseWithPointers versus Graph2ImpreciseDiff.
These output DSLs are the same except ImpreciseWithPointers represents locations with pointers and ImpreciseDiff represents locations with the values of the nodes.
By using the copy mechanism, it would be possible in principle for the ImpreciseDiff model to mimic the ImpreciseWithPointers model.
We suspect the difference in performance comes from the stronger supervision in the ImpreciseWithPointers model---supervision about locations points to exactly the region of the graph that needs to be edited. In the ImpreciseDiff model, the supervision about locations only narrows the location to a set of possible locations that could be copied from.
We also evaluated test accuracy for the DeepDelta and Graph2Diff models that achieve best validation accuracy. DeepDelta test accuracy is 10\% and Graph2Diff is 26\%.
In other words, \emph{Graph2Diff has more
than double the accuracy of DeepDelta,
even though Graph2Diff predicts the change more precisely.}

\begin{figure}
    \centering
\begin{tabular}{|r||c||c|c|c|}
     \hline 
     &
      {\bf Diff}     &
 {\bf \pbox{15cm}{Imprec+PS}}  &
 {\bf Imprec+P} &
 {\bf Imprec} \\
     \hline 
     \hline
     \pbox{20cm}{{\bf Seq2}}
                &
             --- & 
             --- & 
             --- &
             \parbox{1.25cm}{\centering -12.9\%
                          \\ \footnotesize (DeepDelta) }\\ 
             \hline
     \pbox{20cm}{{\bf SeqGraph2}}
                 &
                 -20.3\% & 
                 -13.5\% &
                 -11.4\% &
                 -4.8\% \\
                \hline
     \pbox{20cm}{{\bf Graph2}}
                 &
                 0.0\% & 
                 +3.3\% &
                 +8.0\% &
                 +4.7\% \\
                 \hline
\end{tabular}
\caption{
Absolute sequence-level accuracy difference versus Graph2Diff model. Rows correspond to input representations and columns correspond to output DSLs with increasingly less precision. Only the Diff column contains precise information needed to apply the change unambiguously.
}
    \label{fig:seq_graph_to_abstract_diff_results}
\end{figure}

\subsection{RQ3: How often do incorrect fixes build?}

When deploying the system in practice, we can increase precision by filtering out suggestions 
that do not actually result in a successful build.
In this section we evaluate how effective this filtering step is.
As a first filtering step, we remove all proposed test set fixes from the best Graph2Diff model that do not follow the grammar of our DSL (5\%).
From the remaining incorrect examples, we sample 50 examples and attempt
to build the predicted change.
Of these, 13 (26\%) build successfully.
If we extrapolate these results,
this means that
the model is able to make suggestions for 46\% of the build errors in our data set. Of these, 61\% of the time, the fix exactly matches the one that was eventually suggested by the developer.

\subsection{RQ4: Where is the model successful \& not?}

\subsubsection{Example Correct and Incorrect Fixes} 
To illustrate correct and incorrect predictions, we choose to zoom in on a single diagnostic kind 
(Incompatible types), because it gives a clearer sense of the variety of fix patterns needed to resolve similar errors.
It is also interesting because it is not the most common diagnostic kind (it is fifth),
so this allows exploration of what
the model has learned about the long tail
of potential repairs.
\figref{fig:correct_predictions} shows examples where the model predicted the full output sequence correctly (top) and incorrectly (bottom).
Interestingly, in many of these cases the
fixes seem to depend on details of the relevant APIs.
For example, in the first and fourth correct examples, it generates an extra method call, presumably using the diagnostic message and surrounding context to determine the right method to call.
The second example replaces the type in a declaration, which requires generating a rare token via the copy mechanism.
The third example is a relatively small change in terms of text but takes 17 tokens in the output sequence to generate (see \figref{fig:example_edit_script} (b)).
The third example correctly converts an integer literal to a long literal.
At the bottom, the first example illustrates that one limitation of the approach is understanding the type signatures of rare methods (\texttt{getWidget}). The last example is simply hard to predict without knowing more developer intent.

\begin{figure*}[t]
\begin{tabular}{|C{.35\textwidth} |l|}
\hline
 {\bf Diagnostics} & {\bf Fix} \\
\hline
\hline 
\diagnostics{incompatible types: Builder cannot be converted to WidgetGroup}
& 
\begin{JavaDiff}
-  WidgetGroup widgetGroup = converter.getWidgetGroup();
+  WidgetGroup widgetGroup = converter.getWidgetGroup().build();
\end{JavaDiff}
\\ \hline
\diagnostics{incompatible types: RpcFuture<LongNameResponse> cannot be converted to LongNameResponse}
&
\begin{JavaDiff}
-  LongNameResponse produceLongNameResponseFromX(
+  ListenableFuture<LongNameResponse> produceLongNameResponseFromX(
\end{JavaDiff}
\\ \hline
\diagnostics{incompatible types: int cannot be converted to Long}
&
\begin{JavaDiff}
-  Long jobId = 1;
+  Long jobId = 1L;
 \end{JavaDiff}
\\ \hline
\diagnostics{incompatible types: ListenableFuture<FooResult> cannot be converted to FooResult}
&
\begin{JavaDiff}
-  FooResult x = client.sendFoo(request, protocol);
+  FooResult x = client.sendFoo(request, protocol).get();
\end{JavaDiff}
\\ \hline
\hline
\diagnostics{incompatible types: GetWidgetResponse cannot be converted to Widget}
& 
\begin{JavaDiff}
-  return widget.start().get();
+  return widget.start().get().getWidget();
\end{JavaDiff}
\\ \hline
\diagnostics{incompatible types: FooResponse cannot be converted to Optional<BarResponse>}
& 
\begin{JavaDiff}
-  return LONG_CONSTANT_NAME;
+  return Optional.empty();
\end{JavaDiff}
\\ \hline
\end{tabular}
\caption{Incompatible type error validation examples predicted (Top) correctly and (Bottom) incorrectly.}
\label{fig:correct_predictions}
\end{figure*}

\subsubsection{Accuracy by Diagnostic Kind} 

\begin{figure}[t]
\footnotesize{
\begin{tabular}{|c|l|}
\hline
{\bf Acc} & {\bf First diagnostic kind} \\
\hline
\hline
86\% & \texttt{compiler.err.unreachable.stmt} \\
\hline
69\% & \texttt{compiler.err.cant.assign.val.to.final.var} \\
\hline
45\% & \texttt{compiler.err.unreported.exception.need.to.catch.or.throw} \\
\hline
42\% & \texttt{compiler.err.non-static.cant.be.ref} \\
\hline
33\% & \texttt{compiler.err.cant.resolve} \\
\hline
29\% & \texttt{compiler.misc.inconvertible.types} \\
\hline
29\% & \texttt{compiler.err.var.might.not.have.been.initialized} \\
\hline
20\% & \texttt{compiler.err.except.never.thrown.in.try} \\
\hline
17\% & \texttt{compiler.err.doesnt.exist} \\
\hline
13\% & \texttt{compiler.err.class.public.should.be.in.file} \\
\hline
12\% & \texttt{compiler.err.cant.apply.symbols} \\
\hline
10\% & \texttt{compiler.err.cant.apply.symbol} \\
\hline
9\% & \texttt{compiler.misc.incompatible.upper.lower.bounds} \\
\hline
9\% & \texttt{compiler.err.abstract.cant.be.instantiated} \\
\hline
9\% & \texttt{compiler.err.cant.deref} \\
\hline
6\% & \texttt{compiler.err.does.not.override.abstract} \\
\hline
3\% & \texttt{compiler.err.already.defined} \\
\hline
0\% & \texttt{compiler.err.missing.ret.stmt} \\
\hline
\end{tabular}
}
\caption{
Accuracy by kind of the first diagnostic for those that appeared at least 10 times in validation data.
}
    \label{fig:accuracy_by_kind}
\end{figure}

\figref{fig:accuracy_by_kind} reports accuracy by the kind of the first diagnostic message.
We show results for diagnostic kinds that appear at least 10 times in the validation data.
The model learns to fix many kinds of errors, although there is a clear difference
in the difficulty of different kinds.
For example, the model never correctly fixes a ``missing return statement'' error.
We suspect these fixes are difficult because they are closer to program synthesis, where the
model needs to generate a new line of code that satisfies a variety of type constraints imposed by the context.

\subsubsection{Incorrect Fixes that Build Successfully} 
Finally, we provide three examples from the sampled predictions where the fix is not equivalent to the ground truth, but still builds. The ground truth fix is marked by \texttt{// Ground truth} and the Graph2Diff fix is marked by \texttt{// Graph2Diff fix}. \lstref{lst:manual_fix_one} shows that Graph2Diff is able to import from a different package with the same method name. In fact, in this case, the package imported by the developer is 
deprecated and 
the model's proposed fix is preferred.
\lstref{lst:manual_fix_two} renamed the method differently compared to the ground truth fix. It is one example of suggesting new method name, which has been explored by previous approach \cite{allamanis2015suggesting}. \lstref{lst:manual_fix_three} is one example where the predicted fix is semantically different to the ground truth fix, and it is unlikely that the predicted fix is what the developer intended to do. This is a example of false positive, and it is known as the overfitting problem in the automated program repair community \cite{smith2015cure}.
\begin{JavaDiff}[caption={Import different package}, captionpos=b, basicstyle=\footnotesize, columns=fullflexible, label=lst:manual_fix_one]
+ import static junit.framework.Assert.assertFalse; // Ground truth
+ import static org.junit.Assert.assertFalse; // Graph2Diff fix
\end{JavaDiff}
\begin{JavaDiff}[caption={Changed to a different method name}, captionpos=b, basicstyle=\footnotesize, columns=fullflexible, label=lst:manual_fix_two]
- public void original_method_name() throws Exception
+ public void ground_truth_method_name() throws Exception // Ground truth
+ public void predicted_method_name() throws Exception // Graph2Diff fix
\end{JavaDiff}
\begin{JavaDiff}[caption={Semantically different bug fix}, captionpos=b, basicstyle=\footnotesize, columns=fullflexible, label=lst:manual_fix_three]
- if (id.isEmpty() || Long.parseLong(id).equals(0L))
+ if (id.isEmpty() || Long.valueOf(id).equals(0L)) // Ground truth
+ if (id.isEmpty() || Long.parseLong(id) != 0) // Graph2Diff fix
\end{JavaDiff}

\section{Related Work}
\label{sec:related_work}

\paragraph{Graph Neural Networks to Sequences. }

There has been much recent work on graph neural networks (GNNs) \cite{wu2019comprehensive} but less work on using them to generate sequential outputs.
\citet{li2016gated} map graphs to a sequence of outputs including tokens and pointers to graph nodes. 
The main difference is in the decoder model, which we improve by adding a copy mechanism, feeding back previous outputs (see Supplementary Materials for experiments demonstrating improved performance), and training under weak supervision.
\citet{xu2018graph2seq} present a Graph2Seq model for mapping from graphs to sequences using an attention-based decoder. 
\citet{beck2018graph} develop a graph-to-sequence model with GNNs
and an attention-based decoder. In both cases, there is no copy or pointer mechanism. \citet{song2018graph} develop a model for generating text from Abstract Meaning Representation graphs, which maps from a graph structured input to a sequence output that also has a copy mechanism but not an equivalent of our pointer mechanism.
Finally, there are also some similarities between our model and generative models of graphs \citep{li2018learning,brockschmidt2018generative,you2018graphrnn}, in that these models map from a graph to a sequence of decisions that can include selecting nodes (to determine edges), though it does not appear that either subsumes the other.

\paragraph{Learning Program Repair. }

We refer the reader to \cite{monperrus2018living} for a comprehensive review of program repair. We focus here on the most similar methods.
SequenceR \citep{chen2018sequencer} addresses the problem of program repair based on failing test cases and uses an external fault localization tool to propose buggy lines. A sequence-based neural network with copy mechanism is used to predict a fixed line, using context around the buggy line including the method containing the buggy line and surrounding method signatures. The main differences are that our approach can learn to edit anywhere in the input graph and the use of a graph-structured input representation.

\citet{allamanis2017learning} introduce the Variable Misuse problem and builds a GNN model to predict which variable should be used in a given location. It does not directly address the problem of deciding where to edit, instead relying on an external enumerative strategy.
\citet{vasic2019neural} uses a recurrent neural network with two output pointers to learn to localize and repair Variable Misuse bugs. In our context, the first pointer, which localizes the error, can be thought of as a Tocopo pointer, and the second pointer, which points to the variable that should replace the buggy one, can be thought of as a copy operation in Tocopo. Similar to \cite{li2016gated}, the first predicted pointer is not fed back into the prediction of the second pointer. 

DeepFix is an early work that uses deep learning to fix compilation errors \citep{gupta2017deepfix}. 
They use a seq2seq neural network with attention mechanism to repair a single code line, and multi-line errors can be fixed by multiple passes. 
The input is the whole program, and variables are renamed to reduce the vocabulary.
Then, the model takes the input and predicts the line number along with the bug fix. 
TRACER \citep{ahmed2018compilation} followed the same idea and improved upon DeepFix.
The most significant changes compared to DeepFix are:
\begin{enumerate*}[label={\arabic*)}]
\item The fault localization and patch generation step are separated. TRACER relies on the line number reported by the compiler to localize the bug, while DeepFix outputs a line number and the corresponding bug fix.
\item TRACER's input to the model is much smaller; only the lines surrounding the buggy line are used as input.
\end{enumerate*}
We have taken different design decisions compared to these two approaches. First, we use a copy mechanism to solve the out-of-vocabulary problem instead of renaming the variables, as we believe variable names contain valuable information for understanding the source code. Second, we take into account the whole program as well as the diagnostic information. 
Third, we do not assume that multi-line bugs are independent (e.g.,~row G in \figref{fig:example-fixes}) and fix them in multiple passes. Instead, we use the pointer network to specify different locations and generate all bug fixes simultaneously.

The approach by \cite{mesbah2019fixing} to fix compilation errors is the closest related work, though we focus on all kinds of errors rather than a few common kinds. We compared experimentally and discussed extensively in \secref{sec:deepdelta_experiments}.
Similarly, Getafix also only focuses on a few kinds of errors \cite{scott2019getafix}. They use a clustering algorithm to extract common edit patterns from past bug fixes, and try to apply them on new unseen programs. We have achieved similar results (\figref{fig:accuracy_by_kind}), but on more error types. Our post filtering steps also allow us to obtain a higher precision rate.

\section{Discussion}

We have presented an end-to-end neural network-based approach for localizing and repairing build errors that more than doubles the accuracy of previous work. Evaluation on a large dataset of errors encountered by professional developers doing their day-to-day work shows that the model learns to fix a wide variety of errors.

We hope that the Graph2Tocopo abstraction
is particularly useful for
developing new tools to predict code changes. Graph2Tocopo provides
two ways for tool developers to incorporate their domain expertise into
machine learning models.
First, the input graphs can be expanded to include 
arbitrary information about the program,
including type information
and the results of static and dynamic 
analysis. Once these are added to the graph, the deep learning method can
learn automatically when and how this
information is statistically useful
for predicting fixes.
Second, the edit DSLs can 
be augmented with higher-level
actions that perform more
complex edits that are useful for a specific tool,
such as inserting common idioms,
applying common big fixes,
or even refactoring operations.
Having designed these, the tool developer gains access to state of the art neural network approaches.
The framework generalizes several recent works \citep{allamanis2017learning,chen2018sequencer,vasic2019neural}, and it would be straightforward to express them as Graph2Tocopo problems. 
We are also looking forward to working with tool developers to develop new Graph2Tocopo problems.
We have already benefited from the generality of the Graph2Tocopo abstraction when
running the experiments with different output DSLs in \secref{sec:deepdelta_experiments},
where it was easy to use the same abstraction for a variety of input and output design choices.

More broadly, we hope that \emph{fixing build errors is a stepping stone to related code editing problems:}~there is a natural progression from fixing build errors to other software maintenance tasks that require generating larger code changes.

\ifsubmission\else
\begin{acks}
We thank Petros Maniatis for several valuable discussions and comments on earlier drafts.
\end{acks}
\fi

\bibliographystyle{ACM-Reference-Format}
\bibliography{ms}

\ifsubmission\else
\appendix

\section*{Appendix}

\section*{Table of Contents}
The appendix contains the following:
\begin{itemize}
\item A detailed description of the Graph2Diff neural network architecture and training objective.
\item Formal semantics of Tocopo sequences.
\item Additional experimental details and results.
\end{itemize}

\section{Detailed Description of Graph2Diff Architecture}

In this appendix we describe the Graph2Diff architecture in more detail.
A diagram of the architecture appears in \figref{fig:graph2diff}.

\begin{figure}[t]
\centering
\begin{tabular}{c}
\includegraphics[width=.95\columnwidth]{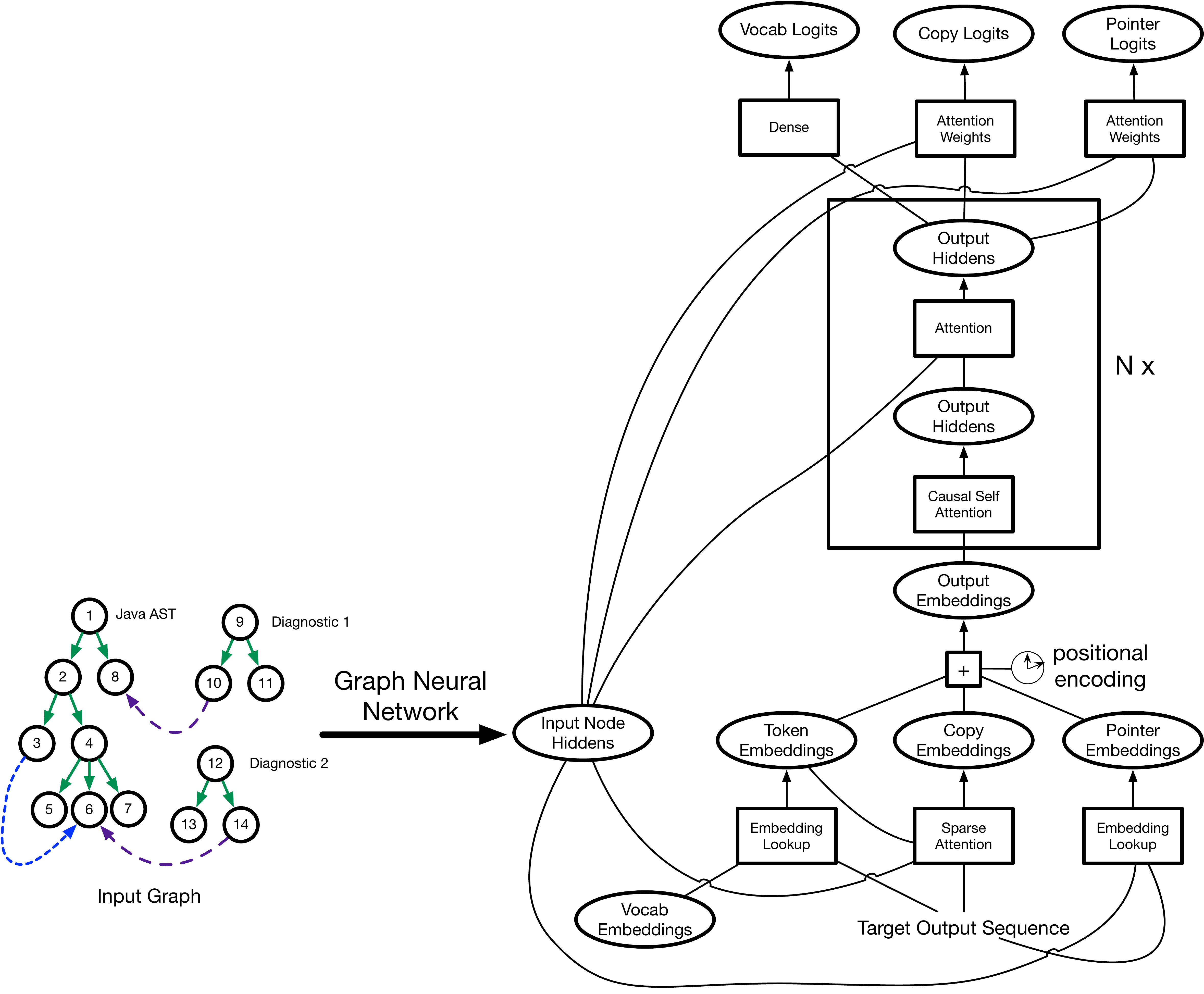}
\end{tabular}
\caption{Graph2Diff model.}
\label{fig:graph2diff}
\end{figure}

\subsection{Background Ops}
We begin by reviewing two operations that we use repeatedly, \emph{sparse attention} \citep{vaswani2017attention,velivckovic2017graph} and \emph{graph propagation} \citep{li2016gated}.

In a sparse attention op, we have $M$ \emph{target} entities, each attending to a subset of $N$ \emph{source} entities. 
Each target entity has a hidden state associated with it, and these hidden states are stacked into an $M \times H$ matrix denoted $\boldsymbol{U}$, where $H$ is the dimension of the hidden states.
Similarly, there is a $N \times H$ matrix of hidden states associated with source entities, denoted $\boldsymbol{V}$.
Each sparse attention op has three dense layers associated with it: 
the \emph{query} layer $f_Q$ transforms $\boldsymbol{U}$ into an $M \times H$ matrix of per-target queries; 
the \emph{key} layer $f_K$ transforms $\boldsymbol{V}$ into an $N \times H$ matrix of per-source keys; and 
the \emph{value} layer $f_V$ transforms $\boldsymbol{V}$ into an $N \times H$ matrix of per-source values. 
Further, the op requires an $M \times N$ sparse binary tensor $\boldsymbol{S}$, where $S_{ji} = 1$ if target $j$ is allowed to attend to source $i$ and 0 otherwise. 
The attention weight from target $j$ to source $i$ is $\alpha_{ji} = \exp \left<f_Q(\boldsymbol{U}_i), f_K(\boldsymbol{V}_j) \right>$, 
and the result of sparse attention is a $M \times H$ matrix where row $j$ is equal to $\frac{1}{Z} \sum_{i : S_{ji}=1} \alpha_{ji} f_V(V_i)$ where $Z = \sum_{i : S_{ji}=1} \alpha_{ji}$.
A special case of sparse attention is \emph{causal self-attention} \cite{vaswani2017attention}, where $\boldsymbol{U}$ and $\boldsymbol{V}$ are both set to be the hidden states associated with output timesteps (i.e., $M$=$N$=\#output steps)
and $S_{ji} = 1$ if $i \le j$ and 0 otherwise.

In a GGNN graph propagation op \citep{li2016gated}, we have a hidden state for each of $N$ nodes in a graph, stacked into a $N \times H$ matrix $\boldsymbol{U}$.
There are $E$ edge types, and each is associated with a sparse tensor $\boldsymbol{S}^{(e)}$ and an associated dense layer $f_e$.
$\boldsymbol{S}^{(e)}_{ji} = 1$ if there is an edge of type $e$ from $i$ to $j$ and 0 otherwise.
The first step is to send messages across all of the edges.
For each node $j$, the incoming messages are defined as $m_j = \sum_e \sum_{i : \boldsymbol{S}^{(e)}_{ji} = 1} f_e(\boldsymbol{U}_i) $.
The second step is to update $\boldsymbol{U}$ by applying a GRU operation \cite{cho2014learning} to each node to get $U'_i = \hbox{GRU}(U_i, m_i)$. 
The result of this update for each node is stacked into a $N \times H$ resulting matrix $\boldsymbol{U}'$.

\subsection{Accommodating Weak Supervision } 
\label{sec:appendix_weak_supervision}
As discussed in \mainsecref{sec:graph2tocopo_learning}, we would like to train our models to maximize the log probability assigned to the set of Tocopo sequences that are reference-equivalent to a given target $\bfs$, which means summing over all valid Tocopo sequences $\boldsymbol{s'} \in \mathcal{I}_{G}(\boldsymbol{s})$.
Our decoder design is motivated by the observation that some architectural choices allow this summation to be computed efficiently.

Note that we can decompose $\mathcal{I}_{G}(\boldsymbol{s})$ into a Cartesian product of per-timestep sets.
Overloading notation, let $\mathcal{I}_{G}(s)$ be the set of Tocopo expressions that evaluate to a target value $s$.
Then $\mathcal{I}_{G}(\boldsymbol{s}) = \mathcal{I}_{G}(s_1) \times \ldots \times
\mathcal{I}_{G}(s_M)$ where $\times$ denotes Cartesian product.
In other words, the set of equivalent Tocopo expressions at time $m$ does not depend on which equivalent Tocopo expression is chosen at other time steps.
As shorthand, let $\mathcal{I}_m = \mathcal{I}_{G}(s_m)$.

We can leverage the above Cartesian product structure to simplify the training of our models with training objective of \maineqref{eq:marginal_likelihood}, but we need to be careful about how the decoder is structured.
Consider defining a model for predicting Tocopo expression $s_m$, and suppose the model has already predicted prefix $s_1 \in \mathcal{I}_1, \ldots, s_{m-1} \in \mathcal{I}_{m-1}$.
If we define predictions in terms of $p(s_m \mid s_1, \ldots, s_{m-1}, \mathcal{G})$, 
where the predicted probability depends on the sequence of \emph{Tocopo expressions} (i.e., \emph{how} the prefix was generated), then the summation in \eqref{eq:marginal_likelihood} requires summing over all valid prefix sequences, which could be exponentially expensive:
\begin{align}
    \log \sum_{\boldsymbol{s} \in \mathcal{I}_1 \times \ldots \times \mathcal{I}_{m}}
    p(s_m \mid s_1, \ldots, s_{m-1}, \mathcal{G}). \label{eq:condition-on-sequence}
\end{align}
However, suppose we define predictions in terms of $p(s_m \mid \\ \mathcal{I}_1, \ldots, \mathcal{I}_{m-1}, \mathcal{G})$, 
where the predicted probability depends on the sequence of equivalence sets.
Then we can write
\begin{align}
\log p(\mathcal{I}_m \mid \mathcal{I}_1, \ldots, \mathcal{I}_{m-1}, \mathcal{G}) =
    \log \sum_{s_m \in \mathcal{I}_{m}}
    p(s_m \mid \mathcal{I}_1, \ldots, \mathcal{I}_{m-1}, \mathcal{G}).
    \label{eq:condition-on-denotation}
\end{align}
In going from Eq.~\ref{eq:condition-on-sequence} to Eq.~\ref{eq:condition-on-denotation}, we lose the ability to assign different probabilities to $p(s_m \mid s_1, \ldots, s_{m-1}, \mathcal{G})$ and $p(s_m \mid s'_1, \ldots, s'_{m-1}, \mathcal{G})$ for two
different prefixes $\boldsymbol{s}$ and $\boldsymbol{s}'$ that are reference-equivalent, but it allows us to perform the marginalization needed for training with weak supervision in linear rather than exponential time.

Our full training objective is thus
\begin{align}
\log p(\mathcal{I}_{G}(\boldsymbol{s}) \mid \mathcal{G}) =
    \sum_m \log \sum_{s_m \in \mathcal{I}_{m}}
    p(s_m \mid \mathcal{I}_1, \ldots, \mathcal{I}_{m-1}, \mathcal{G}).
    \label{eq:condition-on-denotation-full}
\end{align}
From the perspective of the decoder, this means that when feeding back in predictions from previous time steps, we need to feed back the sets $\mathcal{I}_1, \ldots, \mathcal{I}_{m-1}$ and define a neural network model that takes the sets as inputs to the decoder.
This is accomplished via the Output Embedding model described next.

We note that a simpler alternative that preserves efficiency is to not feed back denotation prefixes at all, and instead define objective
\begin{align}
\log p(\mathcal{I}_{G}(\boldsymbol{s}) \mid \mathcal{G}) =
    \sum_m \log \sum_{s_m \in \mathcal{I}_{m}}
    p(s_m \mid m, \mathcal{G}).
\end{align}
In this case, the predictions at the different time steps are conditionally independent of each other given the input graph. 
This is a good choice when there is no uncertainty in the outputs given the inputs, and it is a choice that has been made in previous related work \citep{li2016gated,vasic2019neural}.
However, it cannot represent certain important distributions over output sequences (e.g., .5 probability to make change A at position X and .5 probability to make change B at position Y). We show in \secref{subsec:abGraphStructure} that this alternative leads to significantly worse performance on our problem.

\subsection{Tocopo Decoder}
The decoder predicts the next element of the Tocopo sequence given the input graph and the previous predicted Tocopo elements. 

\paragraph{Output Embedding.} When predicting the next token in the target output sequence, we condition on the previously generated denotations. This section describes how to embed them into hidden states that can be processed by later layers of the decoder. There are three stages of the output embedding.

First, we embed the token associated with each output. For token outputs, this is a standard lookup table into a learnable vector representation per output vocabulary element. Out of vocabulary tokens share a vector. There is no token associated with an input pointer, but we assume they have a special “POINTER” token at this stage (i.e., all pointers share a “POINTER” embedding). 

Second, we incorporate copy information. For each output token, we track which input nodes could have been copied from to generate the token. This information can be represented as a sparse indicator matrix with rows corresponding to output elements and columns corresponding to input nodes, with a 1 entry indicating that the output can be generated by copying from the node. We then perform a sparse attention operation using embeddings from the first output embedding stage as queries and the outputs of the graph encoder as keys and values. The result of this “copy attention” is a weighted average of the final node embeddings of nodes that could be copied from to generate each token, passed through a learnable dense layer. For empty rows of the sparse matrix, the result is the zero vector.

Finally, we incorporate pointer information. This follows similar to the copy embeddings but is simpler because there can be at most one pointer target per output sequence. In terms of the sparse matrix mentioned above, each row has at most a single 1. Thus, we can apply the analogous operation by selecting the node embedding for the column with the 1 and passing it through a different learnable dense layer. For empty rows of the sparse matrix, the result is the zero vector.

The output of each stage is an embedding vector per timestep (which may be the zero vector). The total output embeddings are the sum of the embeddings from the three stages.

\paragraph{Output Propagation.} This stage propagates information from the input graph to the output sequence and between elements of the output sequence. We repeatedly alternate steps of output-to-input attention and output-to-output causal self-attention. The result of each step of the decoder is an updated hidden state for each output step, which is fed as input for the next decoder step. It is initialized to the output of the embedding step above. The result of the propagation stage is the final hidden state for each output step.

Output-to-input attention uses the current output hidden states as keys for a dense attention operation. The keys and values are the final input graph node hidden states. As in \cite{vaswani2017attention}, keys, queries, and values are passed through separate learnable dense layers before performing the attention. The result of attention is a vector for each output step. These are treated as messages and combined with the previous output hidden states using a GRU cell as in GGNNs \citep{li2016gated}. The dense layers and GRU cell parameters are shared across propagation steps. Note that this step allows the output to depend on the entire input graph, even if the input graph has diameter greater than the number of input propagation steps.

Output-to-output attention follows similarly to above but instead of input node hidden states as keys and values, it uses the current output hidden states and masks results so that it is impossible for information about future outputs to influence predictions of previous outputs (i.e., it is causal self-attention \citep{vaswani2017attention}). The output hidden states are updated using a GRU cell as above. The dense layers and GRU cell parameters are shared across propagation steps, but there are separate parameter sets for output-to-input attention and output-to-output attention.

\paragraph{Output Prediction.} Given the result of output propagation, which is a hidden state per output timestep, the final step is to predict a distribution over next outputs. At training time, we simultaneously make predictions for all outputs at once. The output-output propagation ensures that information only flows from previous to future timesteps, so the final hidden state for output step $t$ only includes information about the input graph and outputs up through time $t$. We can thus simply define a mapping from output hidden state $t$ to a distribution over the output $t+1$.

Our approach is to define three output heads. A token head, a copy head, and a pointer head. Letting $H$ be the hidden size, $V$ be the size of the output vocabulary, and $N$ be the number of nodes in the input graph, the token head passes output hidden state $t$ through a dense layer of size $H \times V$. The result is a vector of length $V$ of “vocab logits.” The copy head passes output hidden state $t$ through a dense layer of size $H \times H$ and then computes an inner product of the result with the final representation of each node in the input graph. This gives a size $N$ vector of “copy logits.” The pointer head performs the same operation but with a different dense layer, yielding a size $N$ vector of “pointer logits.” The outputs of the three heads are concatenated together to yield a size $V + 2N$ vector of predictions. We apply a log softmax over the combined vector to get log probabilities for a distribution over next outputs and take the log sum of probabilities associated with correct outputs to compute the training objective.

\section{Formal Semantics of Tocopo Sequences}

We give a formal semantics of what it means for
two Tocopo sequences to refer to the same nodes in the graph.
\begin{table}
\begin{centering}
\fbox{
    \begin{tabular}{rcl}
\denotation{s_1\ldots s_M}{G} 
&=&
\concat{\denotation{s_1 \ldots s_{M-1}}{G},  \denotation{s_M}{G}} \\
\denotation{\texttt{TOKEN}(t)}{G} &=&
(\textsc{Token}, $t$)   \\
\denotation{\texttt{COPY}(i)}{G} &=&
(\textsc{Token}, $v_i$) \\
\denotation{\texttt{INPUT\_POINTER}(i)}{G} &=& (\textsc{InputPointer}, $i$)  \\
    \denotation{\texttt{OUTPUT\_POINTER}(j)}{G} &=&  (\textsc{OutputPointer}, $j$)  \\
    \end{tabular}
} 
\caption{Dereferencing semantics of Tocopo sequences.
Here $s_1 \ldots s_M$ are Tocopo expressions,
 $t \in \calL$ is a token,
 and $i \in 1 \ldots N$ indexes a node in the input graph,
 and $j \geq 0$ is an integer.}\label{tbl:dereferencing_semantics}
\end{centering}
\end{table}

See Table \ref{tbl:dereferencing_semantics}.
Here $\emptyset$ represents an empty sequence,
and $\concat{\cdot}$ adds an element to the end
of a sequence.
To interpret this, notice that many node values
in $G,$ for example, leaf nodes
in the syntax tree, are tokens from $\calL.$
Intuitively, for $\bfs$ and $\bfs'$ to be equivalent,
this means that copy operations are matched
either to copies on nodes that have the same value,
or an equal token from $\calL.$
Intuitively, this semantics dereferences
all of the pointers and copy operations,
but otherwise leaves the Tocopo sequence
essentially unchanged. This is the maximum
amount of semantic interpretation that
we can do without specifying an eDSL.

Now we can define a notion of equivalence.
Our notion of equivalence will
be based on all of 
Two Tocopo sequences $\bfs$ and $\bfs'$ are equivalent with respect to a graph $G$,
which we write $\bfs \topequiv_G \bfs'$, if the sequences have equal derferencing semantics
$\denotation{\bfs}{G} = \denotation{\bfs'}{G}$.
Because it is just checking the location
reference, this notion
equivalence that is generic across eDSLs. 
If $\bfs \topequiv_G \bfs'$, are equivalent Tocopo sequences,
they should be equivalent semantically in any
reasonable eDSL.

\section{Additional Experimental Details and Results}

\subsection{Experimental details}

For these experiments, we randomly split resolutions into 80\% for training, 2500 examples for validation, and 10\% for test.
Unless otherwise specified, all experiments use grid searches over learning rates $\{5e-3, 1e-3, 5e-4, 1e-4\}$, gradient clipping by max norm of $\{.1, 1.0, 10.0\}$, hidden dimension of $\{64, 128\}$ and number of propagation steps $\{1, 2, 4, 8, 12\}$.
Batching follows that of \cite{allamanis2017learning}, packing individual graphs into one large supergraph that contains the individual graphs as separate disconnected components. 
We add graphs to the supergraph as long as the current supergraph has fewer than $20,000$ nodes.
For some of the hyperparameter configurations we exceeded GPU memory, in which case we discarded that configuration.
We allowed training for up to 1M updates, which took on the order of 1 week per run on a single GPU for the larger models.
We report results from the training step and hyperparameter configuration that achieved the best accuracy on the validation data. For Graph2Tocopo models, we use the following vocabulary sizes: 10000 for input graph node values, 1000 for input node types, and 1000 for output vocabulary. For DeepDelta, we use input and output vocab sizes of 30k, as that is the value used by \cite{mesbah2019fixing} and we found it achieve better performance than the smaller vocab sizes used in Graph2Tocopo.

\subsection{Effect of autoregressive feedback}

As discussed in \secref{sec:appendix_weak_supervision}, there is a simpler decoder choice employed by 
\cite{li2016gated} in their latent hiddens model and  \cite{vasic2019neural} that does not feed back previous outputs
when generating a sequence from a graph. We evaluate an ablated form of our model that generates output sequences in the same way,
removing the autoregressive component of our method.
In Figure~\ref{fig:autoregressive_ablation}, we see that feeding in previously predicted outputs to the model is an important component. Leaving it out costs 5-6\% absolute performance (20-26\% relative) degradation.

\begin{figure*}
    \centering
\begin{tabular}{|r||c|c|c|}
     \hline 
     &
      {\bf 2 prop steps}     &
 {\bf 4 prop steps}  &
 {\bf 8 prop steps}  \\
     \hline 
     \hline
     \pbox{20cm}{Prune distance 8 (from above)}
                 &
                 23.8\% & 
                 26.3\% &
                 28.0\%
                 \\
                 \hline
     \pbox{20cm}{Prune distance 8, no feeding back previous \\ 
                 outputs}
                 &
                 17.5\% & 
                 20.8\% &
                 22.3\%
                 \\
                \hline \hline
    Difference (absolute) & 
    -6.3\% &
    -5.5\% &
    -5.7\% \\
    \hline
\end{tabular}
\caption{Effect of removing autoregressive feedback in the decoder.}
    \label{fig:autoregressive_ablation}
\end{figure*}

\subsubsection{Effect of graph structure}
\label{subsec:abGraphStructure}

We ran an experiment removing edge information from the model, preserving only a basic notion of ordering. 
We rendered the diagnostic, build, and Java subgraphs in a linear order via a depth-first pre-order traversal of nodes and removed all edges in the input graphs. We added back edges connecting each node to all nodes within a distance of 10 in the linear ordering of nodes. 
This is meant to mimic a Transformer-style model with local attention \citep{parmar2018image}, working on a linearization of tree structures as input \citep{vinyals2015grammar}.

No hyperparameter setting was able to achieve more than 3\% sequence-level validation accuracy, but the best performing edge-ablated model came from pruning at diameter 1 and doing 12 propagation steps. We speculate that the edges contain critical information for localizing where the edit needs to take place, and this gets lost in the ablation. The diameter 1 problem suffers the least because it has extracted the subgraph around the errors, which makes the localization problem easier (at the cost of not being able to solve as many problems by just looking at that local context).

\fi
\end{document}